\documentclass[sigconf]{acmart}

\usepackage{balance}
\def\BibTeX{{\rm B\kern-.05em{\sc i\kern-.025em b}\kern-.08emT\kern-.1667em\lower.7ex\hbox{E}\kern-.125emX}}

\usepackage{algorithm2e}
\usepackage{amsmath}
\usepackage{graphicx}
\usepackage{booktabs}
\usepackage{tabularx}
\newcolumntype{B}{X}
\newcolumntype{M}{>{\hsize=.3\hsize}X}
\newcolumntype{S}{>{\hsize=.05\hsize}X}
\usepackage{cellspace}
\usepackage{makecell}
\usepackage{todo}
\usepackage{multirow}

\DeclareMathOperator*{\argmin}{argmin}
\DeclareMathOperator*{\argmax}{argmax}

\copyrightyear{2019} 
\acmYear{2019} 
\setcopyright{acmlicensed}
\acmConference[KDD '19]{The 25th ACM SIGKDD Conference on Knowledge Discovery and Data Mining}{August 4--8, 2019}{Anchorage, AK, USA}
\acmBooktitle{The 25th ACM SIGKDD Conference on Knowledge Discovery and Data Mining (KDD '19), August 4--8, 2019, Anchorage, AK, USA}
\acmPrice{15.00}
\acmDOI{10.1145/3292500.3330710}
\acmISBN{978-1-4503-6201-6/19/08}

\acmSubmissionID{ads0801o}

\begin{document}
\title{Unsupervised Clinical Language Translation}

\author{Wei-Hung Weng}
\orcid{0000-0003-2232-0390}
\affiliation{
  \institution{Massachusetts Institute of Technology}
  \city{Cambridge}
  \state{MA 02139, USA}}
\email{ckbjimmy@mit.edu}

\author{Yu-An Chung}
\orcid{0000-0001-9451-7956}
\affiliation{
  \institution{Massachusetts Institute of Technology}
  \city{Cambridge}
  \state{MA 02139, USA}}
\email{andyyuan@mit.edu}

\author{Peter Szolovits}
\affiliation{
  \institution{Massachusetts Institute of Technology}
  \city{Cambridge}
  \state{MA 02139, USA}}
\email{psz@mit.edu}

\renewcommand{\shortauthors}{Weng, Chung, and Szolovits}

\begin{abstract}
As patients' access to their doctors' clinical notes becomes common, translating professional, clinical jargon to layperson-understandable language is essential to improve patient-clinician communication.
Such translation yields better clinical outcomes by enhancing patients' understanding of their own health conditions, and thus improving patients' involvement in their own care.
Existing research has used dictionary-based word replacement or definition insertion to approach the need. 
However, these methods are limited by expert curation, which is hard to scale and has trouble generalizing to unseen datasets that do not share an overlapping vocabulary.
In contrast, we approach the clinical word and sentence translation problem in a completely unsupervised manner.
We show that a framework using representation learning, bilingual dictionary induction and statistical machine translation yields the best precision at 10 of 0.827 on professional-to-consumer word translation, and mean opinion scores of 4.10 and 4.28 out of 5 for clinical correctness and layperson readability, respectively, on sentence translation.
Our fully-unsupervised strategy overcomes the curation problem, and the clinically meaningful evaluation reduces biases from inappropriate evaluators, which are critical in clinical machine learning.
\end{abstract}



\maketitle

\section{Introduction}
Effective patient-clinician communication yields better clinical outcomes by enhancing patients' understanding of their own health conditions and participation in their own care~\cite{ross2003effects}. 
Patient-clinician communication happens not only during in-person clinical visits but also through health records sharing.
However, the records often contain professional jargon and abbreviations that limit their efficacy as a form of communication.
Statistics show that only 12\% of adults are proficient in clinical language, and most consumers can't understand commonly used clinical terms in their health records~\cite{lalor2018comprehenotes}. 
For example, the sentence ``On floor pt found to be hypoxic on O2 4LNC O2 sats 85 \%, CXR c/w pulm edema, she was given 40mg IV x 2, nebs, and put on a NRB with improvement in O2 Sats to 95 \%'' is easy for a trained clinician to understand, yet would not be obvious to typical healthcare consumers, normally patients and their families. 

Clinicians usually provide discharge instructions in consumer-understandable language while discharging patients.
Yet these instructions include very limited information, which does not well represent the patient's clinical status, history, or expectations of disease progression or resolution.
Thus the consumers may not obtain needed information only from these materials. 
To understand more about their clinical conditions for further decision making---for example, seeking a second opinion about treatment plans---it is necessary to dive into the other sections of a discharge summary, which are written in professional language. 
However, without domain knowledge and training, consumers may have a hard time to clearly understand domain-specific details written in professional language.
Such poor understanding can cause anxiety, confusion and fear about unknown domain knowledge~\cite{giardina2011should}, and further result in poor clinical outcomes~\cite{sudore2006limited}
Thus, translating clinical professional to consumer-understandable language is essential to improve clinician-consumer communication and to assist consumers' decision making and awareness of their illness.

Traditionally, clinicians need to specifically write down the consumer-understandable information in the notes to explain the domain-specific knowledge. 
Such a manual approach is acceptable for a small number of cases, but presents a burden for clinicians since the process isn't scalable as patient loads increase. 
An appealing alternative is to perform automated translation. 
Researchers have attempted to map clinical professional to appropriate consumer-understandable words in clinical narratives using an expert-curated dictionary~\cite{zielstorff2003controlled,zeng2006exploring,zeng2007making,kandula2010semantic}, as well as pattern-based mining~\cite{vydiswaran2014mining}. 
However, such methods are either labor-intensive to build dictionaries or raise issues of data reliability and quality, which limit their performance. 

Through advances in representation learning, modern natural language processing (NLP) techniques are able to learn the semantic properties of a language without human supervision not only in the general domain~\cite{mikolov2013distributed,bojanowski2017enriching,peters2018deep,artetxe2018robust,conneau2018word,chung2018unsupervised}, but also in clinical language~\cite{weng2017medical,wang2018comparison,weng2018mapping}.
We aim to advance the state of clinician-patient communication by translating clinical notes to layperson-accessible text. Specifically, we make the following contributions:
\begin{enumerate}
  \item We first design and apply the fully-unsupervised bilingual dictionary induction (BDI) and statistical machine translation (MT) framework for the non-parallel clinical cross-domain (professional-to-consumer) language translation.
  \item We utilize the identical strings in non-parallel corpora written in different clinical languages to serve as anchors to minimize supervision.
  \item We design a clinically meaningful evaluation method which considers both correctness and readability for sentence translation without ground truth reference.
\end{enumerate}

\section{Related Works}

\paragraph{Clinical Professional-Consumer Languages}
To achieve professional-to-consumer language translation in clinical narratives, researchers have attempted to use the dictionary-based~\cite{zielstorff2003controlled,zeng2006exploring,elhadad2007mining,zeng2007making,kandula2010semantic}, and pattern-based mining approaches~\cite{vydiswaran2014mining}. 

Recent studies have mapped clinical narratives to patient-compre\-hensible language using the Unified Medical Language System (UMLS) Metathesaurus combined with the consumer health vocabulary~(CHV) to perform synonym replacement for word translation~\cite{zeng2006exploring,zeng2007making}.
\citet{elhadad2007mining} adopted corpus-driven method and UMLS to construct professional-consumer term pairs for clinical machine translation.
Researchers also utilized external data sources, such as MedlinePlus, Wikipedia, and UMLS, to link professional terms to their definitions for explanation~\cite{polepalli2013improving,chen2018natural}.
However, these dictionary-based approaches have limitations. 
Studies show that expert-curated dictionaries don't include all professional words that are commonly seen in the clinical narratives (e.g. ``lumbar'' is not seen in CHV)~\cite{keselman2008consumer,chen2017ranking}. 
In contrast, the layman terms are not covered well in the UMLS~\cite{elhadad2007mining}. 
Many professional words also don't have corresponding words in consumer language (e.g. ``captopril''), or the translated words are still in professional language (e.g. ``abd'' $\rightarrow$ ``abdomen'').
Such issues limit the utility of dictionaries like CHV to be useful for evaluation but not for training the professional-to-consumer language translation model due to lack of appropriate translation pairs.
Additionally, the definitions of some complex medical concepts in the ontology or dictionary are not self-explanatory.
Consumers may still be confused after translation with unfamiliar definitions. 
Finally, such dictionary curation and expansion are expert-demanding and difficult to scale up. 

\citet{vydiswaran2014mining} applied a pattern-based method on Wiki\-pedia using word frequency with human-defined patterns to explore the relationship between professional and consumer languages. 
The approach is more generalized, yet Wikipedia is not an appropriate proxy for professional language that physicians commonly use in clinical narratives. 
For example, clinical abbreviations such as ``qd'' (once per day) and ``3vd'' (three-vessel coronary artery disease), may not be correctly represented in Wikipedia.
Wikipedia also has great challenges of quality and credibility, even though it is trusted by patients. 
The patterns used to find translation pairs also require human involvement, and the coverage is questionable.
Furthermore, none of the above methods can perform sentence translation that considers the semantics of the context without human supervision, which is a common but critical issue for clinical machine learning. 

\paragraph{Clinical Language Representations}
Recent progress in machine learning has exploited continuous space representations of discrete variables (i.e., tokens in natural language)~\cite{mikolov2013distributed,bojanowski2017enriching,peters2018deep}. 
In the clinical domain, such learned distributed representations (from word to document embeddings) can capture semantic and linguistic properties of tokens in the clinical narratives.
One can directly adopt pre-trained embeddings trained on the general corpus, the biomedical corpus (PubMed, Merck Manuals, Medscape)~\cite{pyysalo2013distributional}, or clinical narratives~\cite{choi2016learning}, for downstream clinical machine learning tasks. 
We can also train the embedding space by fine-tuning the pre-trained model~\cite{hsu2018unsupervised}, or even from scratch---learning the embedding space from one's own corpus~\cite{weng2017medical,weng2018mapping}.
Learned language embedding spaces can also be aligned for cross-domain and cross-modal representation learning by BDI algorithms~\cite{conneau2018word,artetxe2018robust,chung2018unsupervised}.
Researchers have applied such techniques to clinical cross-domain language mapping, as well as medical image-text cross-modal embedding spaces alignment~\cite{weng2018mapping,hsu2018unsupervised}. 
We applied the concepts of the cross-domain embedding spaces alignment to our translation task.

\paragraph{Unsupervised Machine Translation (MT)}
MT has been shown to have near human-level performance with large annotated parallel corpora such as English to French translation. 
However, one big challenge of current MT frameworks is that most language pairs, such as clinical language translation, are low-resource in this sense. 
To make the frameworks more generalizable to low-resource language pairs, it is necessary to develop techniques for fully utilizing monolingual corpora with less bilingual supervision~\cite{lample2018unsupervised,artetxe2018unsupervised,chung2019towards}.

Researchers has developed state-of-the-art neural-based MT frameworks~\cite{lample2018unsupervised,artetxe2018unsupervised}, which first construct a synthetic dictionary using unsupervised BDI~\cite{conneau2018word,artetxe2018robust}.
Then the dictionary is used to initialize the sentence translation. 
Next, the language model is trained and serves as a denoising autoencoder when applied to the encoder-decoder translator to refine the semantics and syntax of the noisy, rudimentary translated sentence~\cite{sutskever2014sequence,bahdanau2014neural,vincent2008extracting}.
Finally, the iterative back-translation is adopted to generate parallel sentence pairs~\cite{sennrich2015improving}. 

Apart from the neural-based approaches, statistical frameworks, such as phrase-based statistical MT (SMT)~\cite{koehn2003statistical}, do not require co-occurrence information to learn the language representations and therefore usually outperform neural-based methods when the dataset and supervision are limited, especially for low-resource language translation. 
In~\cite{lample2018phrase}, they applied the same principles that researchers used in neural-based MT framework to the SMT system and outperformed the neural-based frameworks in some conditions.
We adopted the unsupervised BDI with SMT framework to achieve word and sentence translations.

\section{Methods}
\begin{figure}[htbp]
\centering
\includegraphics[width=1.0\linewidth]{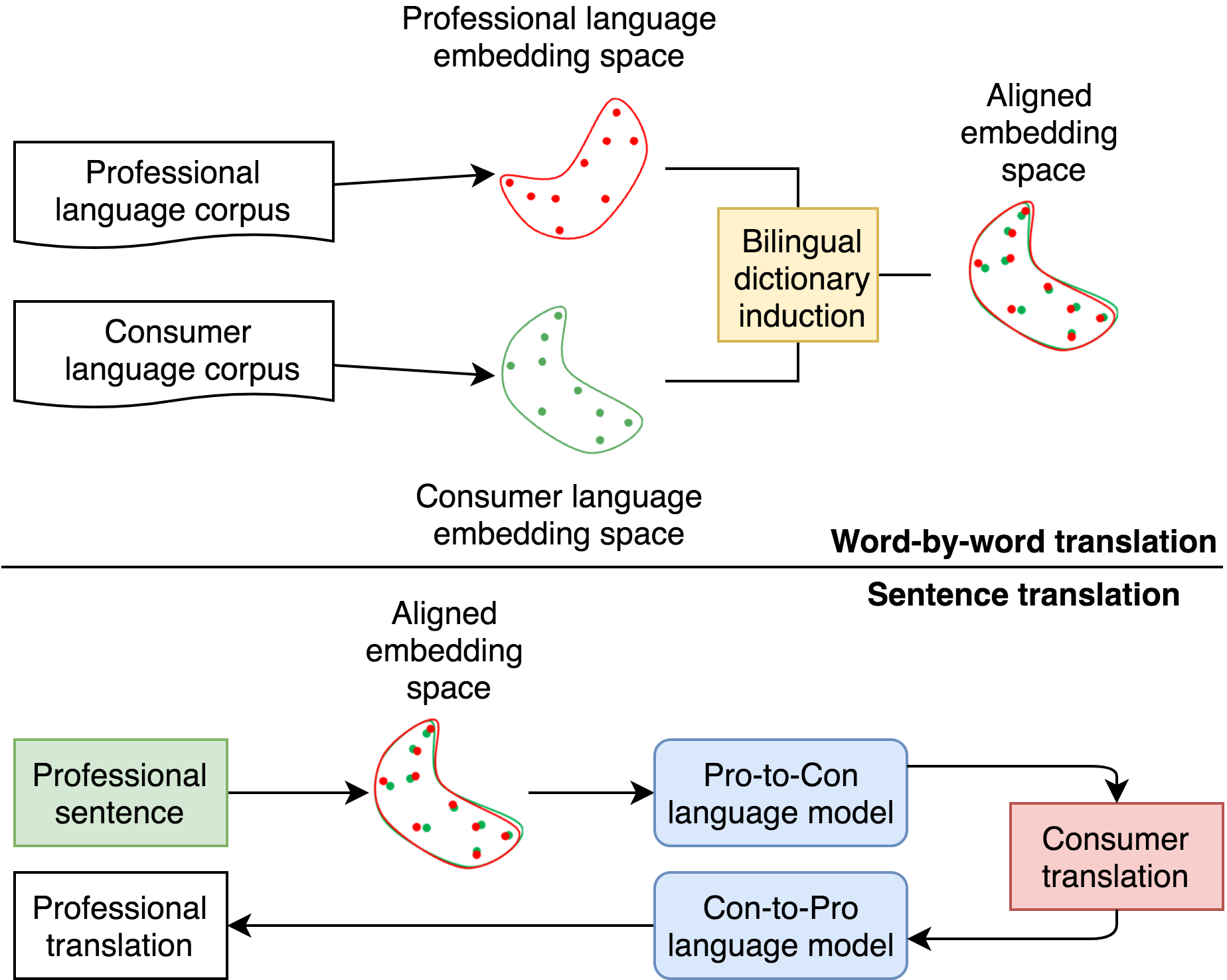} 
\hfill
\vspace{-10pt}
\caption{Overview of our framework. The framework is composed of two steps: (1) word translation through unsupervised word representation learning and bilingual dictionary induction (BDI), and (2) sentence translation, which is initialized by the BDI-aligned word embedding spaces and refined by a statistical language model and back-translation.}
\label{fig:fig1}
\vspace{-10pt}
\end{figure}

The two-step framework is built on several unsupervised techniques for NLP. 
First, we developed a word translation system that translates professional words into consumer-understandable words without supervision. 
Next, we adopted a state-of-the-art statistical MT (SMT) system, which uses language models and back-translation to consider the contextual lexical and syntactic information for better quality of translation. 
The framework follows Figure~\ref{fig:fig1}. 

\subsection{Learning Word Embedding Spaces}
We applied the unsupervised skip-gram algorithm to learn the embedding space of the words that preserve the semantic and linguistic properties~\cite{mikolov2013distributed}. 
The skip-gram model is trained to maximize, for each token $w(n)$ in a corpus, the probability of tokens $\{w_{n-k}, ..., w_{n-1}, w_{n+1}, ..., w_{n+k}\}$ within a window of size $k$ given $w(n)$. 
Word-level representations can also be learned by adding subword information, namely character-level $n-$gram properties that capture more lexical and morphological features in the corpus ~\cite{bojanowski2017enriching}.
We investigated the qualities of learned embedding spaces trained by the skip-gram algorithm with or without subword information.

The assumption of good BDI for translation is that the embedding spaces of source and target languages should be as similar as possible.
Since human languages use similar semantics for similar textual representations~\cite{barone2016towards}, the nearest neighbor graphs derived from word embedding spaces in different languages are likely to be approximately isomorphic.
Thus, it is theoretically possible to align embedding spaces trained by the same algorithm if they have similar shapes of distributions.
To evaluate the similarity between embedding spaces, we compute the eigenvector score between them~\cite{sogaard2018limitations}. 
Higher eigenvector score indicates that the given two embedding spaces are less similar. 
Derived from the eigenvalues of Laplacian matrices, the eigenvector score can be computed as follows: 

\begin{itemize}
    \item Derive the nearest neighbor graphs, $G_1, G_2$, from the learned embedding spaces, then compute $L_1 = D_1 - A_1$ and $L_2 = D_2 - A_2$, where $L_i, D_i, A_i$ are the Laplacian matrices, degree matrices, and adjacency matrices of $G_i$, respectively.
    \item Search for the smallest value of $k$ for each graph such that the sum of largest $k$ Laplacian eigenvalues is smaller than 90\% of the summation of all Laplacian eigenvalues.
    \item Select the smallest $k$ across two graphs and compute the squared differences, which is the eigenvector score, between the largest~$k$ eigenvalues in two Laplacian matrices.
\end{itemize}

\subsection{Bilingual Dictionary Induction for Word Translation}
Unsupervised BDI algorithms can be applied to learn a mapping dictionary for alignment of embedding spaces.
We investigated two state-of-the-art unsupervised BDI methods: (1) iterative Procrustes process (\texttt{MUSE})~\cite{conneau2018word} and (2) self-learning (\texttt{VecMap})~\cite{artetxe2018robust}. 
The goal of alignment is to learn a linear mapping matrix~$W$. 
To minimize supervision, we did not use any mapping dictionaries, such as CHV, but leveraged the characteristics of two English corpora to use identical strings in two corpora to build a synthetic seed dictionary. 

\paragraph{Using Anchors}
The identical strings served as anchors to learn $W$ with \texttt{MUSE} or \texttt{VecMap}.
\texttt{MUSE} adopted the technique of the Procrustes process, which is a linear transformation. 
Assuming that we have the $x$-word, $d$-dimensional professional language embedding~$\mathcal{P} = \{p_{1}, p_{2}, \ldots, p_{x}\}\subseteq \mathbb{R}^{d}$ and the $y$-word, $d$-dimension consumer language embedding~$\mathcal{C} = \{c_{1}, c_{2}, \ldots, c_{y}\}\subseteq \mathbb{R}^{d}$. 
We used $k$ anchors to build the synthetic mapping dictionary and learn~$W$ between the two embedding spaces, such that~$p_{i}\in \mathcal{P}$ maps to the appropriate~$c_{j}\in \mathcal{C}$ without supervision. 
Then we have:
$W^{\star} = \argmin_{W\in \mathbb{R}^{d \times d}} \|WX - Y\|^{2}$
, where~$X = \{x_1, x_2, \ldots, x_k\} \subseteq \mathbb{R}^{d}$ and~$Y = \{y_1, y_2, \ldots, y_k\} \subseteq \mathbb{R}^{d}$ are two aligned matrices of size~$d \times k$ formed by~$k$-word embeddings selected from~$\mathcal{P}$ and~$\mathcal{C}$.

An orthogonality constraint is added on~$W$, where the above equation will turn into the Procrustes problem that can be solved by singular value decomposition (SVD) with a closed form solution~\cite{xing2015normalized}:
\begin{equation}
\resizebox{0.92\columnwidth}{!}{$
W^{\star} = \argmin_{W\in \mathbb{R}^{d \times d}} \|WX - Y\|^{2} = UV^T, \text{ where } U \Sigma V^T = \textsf{SVD}(YX^T)
$}
\end{equation}
The aligned output of the professional language input~$p_i$, i.e. the best translation $c_j = \argmax_{c_{j}\in \mathcal{C}}\cos(Wp_{i}, c_{j})$.

For the \texttt{VecMap} self-learning method, the idea includes two steps~\cite{artetxe2018robust}.
First, using a dictionary $D_{ij}$ to learn the mappings $W_X, W_Y$ that will transform both $X$ and $Y$ to maximize the similarity for the given dictionary as follows:
\begin{equation}
\begin{split}
&\argmax_{W_X, W_Y} \sum_i \sum_j D_{ij} (W_X x_i \cdot W_Y y_j) \\
\end{split}
\end{equation}
where the optimal result is given by $W_X \Sigma W_Y^T = \mathsf{SVD}(X^T D Y)$. 
We again utilized the identical strings to build the initial dictionary.
Symmetric re-weighting of $X, Y$ is applied before and after SVD~\cite{artetxe2018generalizing}. 

Next, we use $W_X, W_Y$ to bidirectionally compute the updated dictionary over the similarity matrix of the mapped embedding, $XW_XW_Y^TY^T$.
The values in the updated dictionary are filled using Cross-Domain Similarity Local Scaling (CSLS), where the value equals 1 if translation $y_j = \argmax_{y_j \in Y} (W_X x_i \cdot W_Y y_j)$, else equals zero.
The above two steps are trained iteratively until convergence. 

\paragraph{Without Anchors}
We adopted adversarial learning for the configurations if identical strings were not used. 
We first learned an approximate proxy for $W$ using a generative adversarial network (GAN) to make~$\mathcal{P}$ and~$\mathcal{C}$ indistinguishable, then refined by the iterative Procrustes process to build the synthetic dictionary~\cite{conneau2018word,goodfellow2014generative}.

In adversarial learning, the discriminator aims to discriminate between elements randomly sampled from~$W\mathcal{P} = \{Wp_{1}, Wp_{2}, \ldots, Wp_{x}\}$ and~$\mathcal{C}$.
The generator, $W$, is trained to prevent the discriminator from making an accurate prediction.
Given $W$, the discriminator parameterized by~$\theta_{D}$ tries to minimize the following objective function ($\text{Pro}=1$ indicates that it is in the professional language):
\begin{equation}
\begin{split}
\resizebox{0.92\columnwidth}{!}{$
  \mathcal{L}_{D}(\theta_{D}|W) = -\frac{1}{x}\sum_{i = 1}^{x}\log \mathbb{P}_{\theta_{D}}(\text{Pro} = 1 | Wp_{i}) 
  - \frac{1}{y}\sum_{j = 1}^{y}\log \mathbb{P}_{\theta_{D}}(\text{Pro} = 0 | c_{j})
$}
\end{split}
\end{equation}
Instead,~$W$ minimizes the following objective function to fool the discriminator:
\begin{equation}
\begin{split}
\resizebox{0.92\columnwidth}{!}{$
  \mathcal{L}_{W}(W|\theta_{D}) = -\frac{1}{x}\sum_{i = 1}^{x}\log \mathbb{P}_{\theta_{D}}(\text{Pro} = 0 | Wp_{i})
  - \frac{1}{y}\sum_{j = 1}^{y}\log \mathbb{P}_{\theta_{D}}(\text{Pro} = 1 | c_{j})
$}
\end{split}
\end{equation}
The optimizations are executed iteratively to minimize~$\mathcal{L}_{D}$ and~$\mathcal{L}_{W}$ until convergence.

We performed nearest neighbors word retrieval using CSLS instead of simple nearest neighbor (NN). 
The purpose of using CSLS is to reduce the problem of ``hubness,'' that a data point tends to be nearest neighbors of many points in a high-dimensional space due to the asymmetric property of NN~\cite{conneau2018word,artetxe2018robust,dinu2015improving}.
\begin{equation}
\resizebox{0.92\columnwidth}{!}{$
\begin{split}
\textsf{CSLS}(W\mathcal{P}, \mathcal{C}) &= 2 \cos (W\mathcal{P}, \mathcal{C}) \\
&- \frac{1}{k} \sum_{c_j \in \textsf{NN}_Y (Wp_i)} \cos(Wp_i, c_j) 
- \frac{1}{k} \sum_{Wp_i \in \textsf{NN}_X (c_j)} \cos(Wp_i, c_j)
\end{split}
$}
\end{equation}

Word translation is done using BDI algorithms by a series of linear transformations. 
However, language translation requires not only the word semantics, but also the semantic and syntactic correctness at the sentence level.
For instance, the ideal translation is not the nearest target word but synonyms or other close words with morphological variants. 
Further refinement is therefore necessary for sentence translation.

\subsection{Sentence Translation} 
The unsupervised phrase-based SMT includes three critical steps:
\begin{enumerate}
    \item Careful initialization with word translation, 
    \item Language models for denoising,
    \item Back-translation to generate parallel data iteratively.
\end{enumerate}

We initialized the sentence translation with the aligned word embedding spaces trained by unsupervised word representation learning and BDI algorithms. 

To translate the word in professional language $p_i$ to the word in consumer language $c_j$, the SMT scores $c_j$ where $\argmax_{c_j} \mathbb{P}(c_j | p_i) = \argmax_{c_j} \mathbb{P}(p_i | c_j) \mathbb{P}(c_j)$. 
The $\mathbb{P}(p_i | c_j)$ is derived from the phrase tables and $\mathbb{P}(c_j)$ is from a language model~\cite{lample2018phrase}. 
We used the mapping dictionary generated by the BDI algorithm as the initial phrase (word) table to compute the softmax scores, $\mathbb{P}(c_j | p_i)$, of the translation of a source word, where
\begin{equation}
\mathbb{P}(c_j | p_i) = \frac{\exp(T^{-1}\cos[W  emb(p_i), emb(c_j)])}{\sum_k \exp(T^{-1}\cos[W emb(p_i), emb(c_k)])}
\end{equation}
where $emb(x)$ is the embedding of word $x$, cos is the cosine similarity, and $T$ is a hyperparameter for tuning the peakness of the distribution. 
We then learned smoothed n-gram language models using KenLM for both professional and consumer corpora~\cite{heafield2011kenlm}.

Next, we used the initial phrase table and language models mentioned above to construct the first rudimentary SMT system to translate the professional sentence into consumer language. 
Once we got the translated sentences, we were able to train a backward SMT from target to source language (back-translation) by learning new phrase tables and language models.
We can therefore generate new sentences and phrase tables to update language models in two directions, back and forth, for many iterations.

\section{Data}
Data was collected from the Medical Information Mart for Intensive Care III (MIMIC-III) database~\cite{johnson2016mimic}, containing de-identified data on 58,976 ICU patient admissions to the Beth Israel Deaconess Medical Center (BIDMC), a large, tertiary care medical center in Boston, Massachusetts, USA. 
We extracted 59,654 free-text discharge summaries from MIMIC-III. 
Clinical notes usually have many sections.
Among all sections, we selected the ``History of present illness'' and ``Brief hospital course'' sections to represent the content with professional jargon because these sections are usually the most narrative components with thoughts and reasoning for the communication between clinicians.
In contrast, ``Discharge instruction'' and ``Followup instruction'' sections are written in consumer language for patients and their families. 
We omit other sections since they are usually not written in natural language but only lists of jargon terms, such as a list of medications or diagnoses.

Although the professional and consumer corpora are both from MIMIC-III, their content are not parallel.
However, we expect that there are identical strings across two corpora since both of them are written in English. 
We utilized 4,605 overlapping English terms as anchors to create a seed dictionary in BDI to minimize supervision.

We also collected additional consumer language data from the English version of the MedlinePlus corpus\footnote{\texttt{https://medlineplus.gov/xml.html}}. 
MedlinePlus is the patient and family-oriented information produced by the National Library of Medicine. 
The corpus is about diseases, conditions, and wellness issues and written in consumer understandable language. 
We investigated whether the addition of the MedlinePlus corpus enhances the quality of BDI.

The statistics of the corpora used are shown in Table~\ref{tab:table1}.
\begin{table}[htbp]
\begin{tabular}{ccc}
\toprule
Corpus & \#Sentence & \#Vocabulary \\ 
\midrule
MIMIC-professional language & 443585 & 19618 \\
MIMIC-consumer language & 73349 & 5264 \\
MIMIC-consumer + MedlinePlus & 87295 & 6871 \\
\bottomrule
\end{tabular}
\caption{The detailed statistics of the corpora.}
\label{tab:table1}
\vspace{-20pt}
\end{table}

For data preprocessing, we removed all personal health information placeholders in the MIMIC corpora, then applied the Stanford CoreNLP toolkit and Natural Language Toolkit (NLTK) to perform document sectioning and sentence fragmentation~\cite{manning2014stanford}.

To build the language models for sentence translation, we experimented with using either the MIMIC-consumer corpus or a general corpus, for which we used all sentences from the WMT English News Crawl corpora from years 2007 through 2010, which include 38,214,274 sentences from extracted online news publications.

\section{Experiments}
In this study, we consider MT in two parts: (1) word translation, and (2) sentence translation. We define the tasks and overview of evaluations in this section. For details of model architectures, training settings and evaluations, please refer to the Supplemental Material~\ref{app:1}.

\subsection{Word Translation}
We adopted the skip-gram algorithm to learn word embeddings.
We investigated (1) whether adopting subword information to train word embedding spaces is useful, (2) if different BDI methods (\texttt{MUSE} or \texttt{VecMap}) matter, (3) whether integrating MedlinePlus to augment the consumer corpus is helpful, (4) what dimensionality of word embedding spaces is optimal. 

We evaluated the quality of word translation through nearest neighbor words retrieval. 
Two evaluations were performed.
First, we used a list of 101 professional-consumer word pairs developed by trained clinicians based on their commonly-used professional words. 
The word pairs list was further reviewed and approved by non-professionals with expert explanations. 
Several examples of the ground truth pairs include: (bicarbonate, soda), (glucose, sugar), (a-fib, fibrillation), (cr, creatinine), (qd, once/day). 
Since 14 out of 101 evaluation ground truth pairs do not appear in the training corpora, we used the matched 87 pairs for all quantitative evaluations. 
We also evaluated our method on CHV pairs, which includes 17,773 unique word pairs.
We chose the configurations and parameters for sentence translation based on the results of these two evaluations.

We show the performance by computing precision at $k$, where we used CSLS to query the nearest $k$ words ($k=1, 5, 10$) in the consumer language embedding space using the words in the aligned professional language embedding space.

\subsection{Sentence Translation}
The goal of sentence translation is to translate the sentence in the professional language domain into a sentence in the consumer language domain. We applied the SMT framework and examined the quality of translation by considering whether (1) subword information, (2) anchors for BDI, and (3) language model trained on specific or general corpus, are helpful. 

We adopted Moses, a widely-used statistical MT engine that is used to train statistical translation models~\cite{koehn2007moses}. 
We used the supervised, dictionary-based CHV professional-to-consumer word mapping and replacement as the strong baseline since the replacement mainly preserves clinical correctness.
The Wikipedia pattern-based approach was not considered due to the issues of credibility and quality.
Detailed configurations of SMT are shown in Table~\ref{tab:table2}.
\begin{table}[htbp]
\resizebox{\columnwidth}{!}{
\begin{tabular}{cccc}
\toprule
\textbf{Configuration} & \textbf{Word embedding} & \textbf{Anchors} & \textbf{Language model} \\
\midrule
\textit{A} & 100d with subword & Y & WMT \\
\textit{B} & 100d with subword & Y & MIMIC-consumer \\
\textit{C} & 1000d with subword + augmentation & Y & WMT \\
\textit{D} & 1000d with subword + augmentation & Y & MIMIC-consumer \\
\textit{E} & 300d w/o subword & Y & WMT \\
\textit{F} & 300d w/o subword & Y & MIMIC-consumer \\
\textit{N} & 300d w/o subword & N & WMT \\
\bottomrule
\end{tabular}
}
\caption{Configurations of statistical MT (SMT) for sentence translation.}
\label{tab:table2}
\vspace{-20pt}
\end{table}

Since there is no ground truth reference for clinical professional-to-consumer sentence translation, using standard quantitative metrics such as BLEU or CIDEr score is not possible. 
Previously, researchers asked either clinical experts~\cite{zeng2007making,chen2018natural}, or crowd-sourced Amazon Mechanical Turks (AMT) to score outputs or provide feedback on readability of mapped terms~\cite{lalor2018comprehenotes}.
Instead, we not only invited non-clinicians to score and provide their comments on readability, but we also asked clinicians to evaluate the correctness of the translations before reaching out the non-clinicians to evaluate readability.
We adopted the two-step evaluation because clinical correctness is critical but hard to evaluate by non-clinicians; and by contrast, judgment of readability may be biased for clinicians.

We recruited 20 evaluators---10 clinical professionals and 10 non-clinicians.
For each evaluator, we randomly assigned 20 sentence sets. 
Each set includes the professional sentence (PRO), the translated sentence using configuration $A, B$ (or $C, D$), $E, F, N$, and CHV baseline. 
We asked evaluators to score the translated sentences.

We adopted the mean opinion score (MOS) to evaluate the quality of translation. 
Our MOS evaluation includes two steps (Figure~\ref{fig:figS1}). 
We first asked the clinicians to provide the correctness score of each translated sentence, score ranging from 1 (the worst) to 5 (the best). 
If the correctness score of the translated sentence is less than 4, the sentences will be discarded and not further scored by both professionals and non-professionals since the sentence is not clinically correct, as judged by professionals, and thus meaningless to score for readability.
Otherwise, the sentences will be assigned to both clinicians and non-clinicians for readability scoring. 
The final MOS were computed by averaging all given valid scores.
For the criteria and examples of correctness and readability scoring, please refer to the Supplemental Material~\ref{app:2}.

\section{Results and Discussions}

\subsection{Word Translation}
\paragraph{Bilingual Dictionary Induction Algorithm and Data Augmentation}
In Table~\ref{tab:table3}, we demonstrate that \texttt{MUSE} generally outperforms \texttt{VecMap}. 
We also identified a trend that the performance is better when consumer corpus augmentation was not used.
The only exception was when we applied the corpus augmentation to subword embeddings while evaluating on CHV pairs.

The nature of MedlinePlus texts is very different from clinical narratives since the former are articles for general patient education whereas the latter are more specific to individual cases and colloquial. 
It is highly likely that MedlinePlus and MIMIC-consumer corpora have very different data distributions and therefore affect the quality of BDI.
This also yields inferior performance when we did evaluation on the clinician-designed word pairs since they are also in clinical narrative rather than literature style.
In contrast, CHV covers many morphologically similar words that are shown in the literature but rare in clinical narratives, which results in better performance while using subword embeddings with MedlinePlus augmentation while evaluating on CHV pairs.

By computing eigenvector score, we found that no augmentation yielded smaller eigenvector score (smaller difference between embedding spaces) than with augmentation. 
Eigenvector scores increase from 0.035 to 0.177 (without subword information), and 0.144 to 0.501 (with subword information), after consumer corpus augmentation, which also indicates that adding MedlinePlus yields harder BDI.
Since the embedding space similarity is higher without augmentation, \texttt{MUSE} can perform well in such conditions, as mentioned in previous literature~\cite{artetxe2018robust}.

\begin{table}[htbp]
\resizebox{\columnwidth}{!}{
\begin{tabular}{ccccccc}
\toprule
\textbf{} & \textbf{} & \textbf{} & \multicolumn{2}{c}{Without subword} & \multicolumn{2}{c}{With subword} \\ \cline{4-7} 
 &  &  & \texttt{MUSE} & \texttt{VecMap} & \texttt{MUSE} & \texttt{VecMap} \\ \hline
\multicolumn{1}{c}{\multirow{6}{*}{Clinician}} & \multicolumn{1}{c}{\multirow{2}{*}{P@1}} & aug(+) & 15.61 (3.57) & 15.73 (2.40) & 15.49 (2.70) & 15.24 (1.75) \\
\multicolumn{1}{c}{} & \multicolumn{1}{c}{} & aug(-) & \textbf{17.78 (3.25)} & 13.46 (1.69) & \textbf{20.62 (2.85)} & 20.37 (4.09) \\ \cline{2-7} 
\multicolumn{1}{c}{} & \multicolumn{1}{c}{\multirow{2}{*}{P@5}} & aug(+) & 39.02 (2.37) & 37.19 (4.61) & 42.56 (4.82) & 40.97 (3.65) \\
\multicolumn{1}{c}{} & \multicolumn{1}{c}{} & aug(-) & \textbf{42.84 (2.25)} & 36.71 (3.76) & \textbf{48.15 (3.99)} & 42.10 (2.88) \\ \cline{2-7} 
\multicolumn{1}{c}{} & \multicolumn{1}{c}{\multirow{2}{*}{P@10}} & aug(+) & 46.95 (3.69) & 45.85 (4.57) & 54.76 (3.61) & 52.56 (2.11) \\
\multicolumn{1}{c}{} & \multicolumn{1}{c}{} & aug(-) & \textbf{53.86 (5.75)} & 47.78 (5.73) & \textbf{58.27 (3.53)} & 49.51 (3.37) \\ \hline
\multicolumn{1}{c}{\multirow{6}{*}{CHV}} & \multicolumn{1}{c}{\multirow{2}{*}{P@1}} & aug(+) & \multicolumn{1}{l}{17.94 (2.54)} & \multicolumn{1}{l}{13.51 (1.97)} & \multicolumn{1}{l}{\textbf{22.11 (2.83)}} & \multicolumn{1}{l}{18.51 (1.54)} \\
\multicolumn{1}{c}{} & \multicolumn{1}{c}{} & aug(-) & \multicolumn{1}{l}{\textbf{18.26 (2.72)}} & \multicolumn{1}{l}{13.09 (1.95)} & \multicolumn{1}{l}{21.29 (2.29)} & \multicolumn{1}{l}{16.70 (1.46)} \\ \cline{2-7} 
\multicolumn{1}{c}{} & \multicolumn{1}{c}{\multirow{2}{*}{P@5}} & aug(+) & \multicolumn{1}{l}{36.04 (4.01)} & \multicolumn{1}{l}{29.40 (2.38)} & \multicolumn{1}{l}{44.06 (3.61)} & \multicolumn{1}{l}{38.82 (2.15)} \\
\multicolumn{1}{c}{} & \multicolumn{1}{c}{} & aug(-) & \multicolumn{1}{l}{\textbf{37.30 (3.87)}} & \multicolumn{1}{l}{29.39 (2.54)} & \multicolumn{1}{l}{\textbf{44.92 (3.07)}} & \multicolumn{1}{l}{37.01 (3.01)} \\ \cline{2-7} 
\multicolumn{1}{c}{} & \multicolumn{1}{c}{\multirow{2}{*}{P@10}} & aug(+) & \multicolumn{1}{l}{43.82 (4.15)} & \multicolumn{1}{l}{36.36 (2.61)} & \multicolumn{1}{l}{\textbf{53.73 (3.98)}} & \multicolumn{1}{l}{48.65 (3.65)} \\
\multicolumn{1}{c}{} & \multicolumn{1}{c}{} & aug(-) & \multicolumn{1}{l}{\textbf{45.21 (3.59)}} & \multicolumn{1}{l}{37.99 (3.16)} & \multicolumn{1}{l}{53.42 (4.12)} & \multicolumn{1}{l}{47.27 (3.49)} \\
\bottomrule
\end{tabular}
}
\caption{Performance of nearest neighbors retrieval using CSLS. Comparison between unsupervised Procrustes process (\texttt{MUSE}) and self-learning (\texttt{VecMap}), with or without augmented corpus (MedlinePlus) on clinician-designed pairs evaluation and CHV pairs evaluation. The word embeddings are trained by the fastText skip-gram. For subword information, we considered bigram to 5-gram. We chose a 100-dimensional parameterization, which is common for investigating BDI algorithms and data augmentation. 
The values reported are precision at $k$ ($P@k$) $\times 100$ with standard deviation. Precision at 1 is equivalent to accuracy. Boldface values are the best combination of BDI and augmentation with or without subword information.}
\label{tab:table3}
\vspace{-20pt}
\end{table}

\paragraph{Subword Information and Dimensionality}
Next we searched for the ideal dimensionality beyond the common parameterization.
We used \texttt{MUSE} without MedlinPlus for the following experiments except for the embeddings with subword information evaluated on CHV pairs, for which we augmented the consumer corpus with MedlinePlus.
For clinician-designed pairs, we found that the embeddings enriched with subword information have slightly superior performance to those without using subword information when the embedding space dimension is smaller (Table~\ref{tab:table4}). 
However, embeddings without subword information yield better performance with higher embedding space dimensionality.
In CHV pairs, embeddings with subword information yield superior performance to those without subword information.
Such a finding is reasonable since there are many morphologically similar translation pairs in CHV pairs.
For example, ``asphyxiation'' $\rightarrow$ ``asphyxia''.
They also yield better performance with higher embedding space dimensionality.

The optimal embedding space dimensionality for the subword enriched embedding is 100 dimensions in clinician-designed evaluation pairs, and 1000 dimensions (with MedlinePlus) in CHV pairs evaluation. For the embeddings without subword information, 300 dimensions usually yields better performance.

\begin{table}[htbp]
\resizebox{\columnwidth}{!}{
\begin{tabular}{ccccccc}
\toprule
\textbf{} & \multicolumn{3}{c}{Clinician-designed} & \multicolumn{3}{c}{CHV} \\ \cline{2-7} 
 & Dim & No subword & Subword & Dim & No subword & Subword \\ \hline
\multirow{5}{*}{P@1} & 50 & 12.22 (2.50) & 12.84 (2.19) & 100 & 18.26 (3.02) & 22.11 (1.86) \\
 & 100 & 17.78 (3.25) & \textbf{20.62 (2.85)} & 200 & 20.31 (2.57) & 35.54 (2.53) \\
 & 200 & 21.11 (3.37) & 19.26 (2.92) & 300 & \textbf{22.75 (2.98)} & 46.68 (3.77) \\
 & 300 & \textbf{20.62 (2.80)} & 14.69 (2.05) & 500 & 20.12 (2.12) & 53.89 (4.13) \\
 & 500 & 20.37 (1.77) & 12.96 (1.57) & 1000 & 20.80 (1.95) & \textbf{54.55 (4.52)} \\ \hline
\multirow{5}{*}{P@5} & 50 & 34.44 (3.21) & 38.89 (3.78) & 100 & 37.30 (3.43) & 44.06 (3.81) \\
 & 100 & 42.84 (2.25) & 48.15 (3.99) & 200 & 39.16 (3.67) & 61.34 (4.59) \\
 & 200 & 48.62 (7.11) & \textbf{49.51 (2.57)} & 300 & 41.31 (3.61) & 70.35 (5.02) \\
 & 300 & 48.08 (3.49) & 46.91 (2.54) & 500 & \textbf{42.87 (4.01)} & 77.81 (5.96) \\
 & 500 & \textbf{51.21 (2.14)} & 43.83 (1.95) & 1000 & 40.82 (3.54) & \textbf{76.99 (5.72)} \\ \hline
\multirow{5}{*}{P@10} & 50 & 44.07 (1.75) & 47.41 (3.83) & 100 & 45.20 (3.49) & 53.73 (4.17) \\
 & 100 & 53.86 (5.75) & \textbf{58.27 (3.53)} & 200 & 49.80 (3.28) & 70.35 (5.10) \\
 & 200 & 59.75 (2.19) & 57.78 (4.86) & 300 & 50.49 (4.02) & 76.49 (5.58) \\
 & 300 & \textbf{60.25 (3.63)} & 55.54 (3.62) & 500 & \textbf{51.27 (3.85)} & 82.64 (6.10) \\
 & 500 & 59.01 (2.08) & 53.95 (3.08) & 1000 & 49.71 (3.16) & \textbf{82.71 (5.76)} \\
 \bottomrule
\end{tabular}
}
\caption{Performance of word translation using iterative Procrustes process (\texttt{MUSE}) on clinician-designed pairs and CHV pairs evaluation. The word embeddings are trained by the fastText skip-gram algorithm. For subword information, we considered bigram to 5-gram. The values reported are precision at $k$ ($P@k$) $\times 100$ with standard deviation. Precision at 1 is equivalent to accuracy.}
\label{tab:table4}
\vspace{-20pt}
\end{table}

\paragraph{Qualitative Evaluation}
In Table~\ref{tab:table5}, we demonstrate that the BDI-learned mapping dictionaries are clinically meaningful through CSLS nearest neighbors retrieval. 
Four aligned embedding spaces (dictionaries), including the 100-dimension subword embedding, 300-dimension word embedding, 1000-dimension subword embedding with MedlinePlus augmentation, and 300-dimension word embedding without using anchors, are evaluated qualitatively.
We retrieved the nearest top-5 neighbors in consumer language from the aligned embedding spaces using 12 professional words as queries.
To ensure diversity, 3 anatomy-related, 3 disease-related, 2 procedure-related, 2 lab-related, and 2 medication-related professional words were used for querying.
We found that the appropriate translations were shown in top-5 neighbors in most cases.

\begin{table*}[htbp]
\centering
\resizebox{\textwidth}{!}{
\begin{tabular}{ccccccccccccc}
\toprule
\multicolumn{1}{c}{\textbf{Rank}} & \textbf{cr} & \textbf{cxr} & \textbf{ekg} & \textbf{hepatic} & \textbf{malignancy} & \textbf{mi} & \textbf{na} & \textbf{ophthalmology} & \textbf{qd} & \textbf{renal} & \textbf{sob} & \textbf{vancomycin} \\ \hline
\multicolumn{13}{c}{\textit{Embeddings with Subword Information, with Anchors (Clinician-designed Word Pairs)}} \\ \hline
\multicolumn{1}{c}{1} & kidney & \textit{x-ray} & echocardiogram & portal & \textit{cancer} & stenosis & bun & ob/gyn & qday & \textit{kidney} & shortnes & \textit{antibiotic} \\
\multicolumn{1}{c}{2} & \textit{creatinine} & \textit{xray} & echocardiograms & biliary & carcinoma & \textit{attack} & liter & ob-gyn & dailyname & \textit{kidneys} & shortness & ceftriaxone \\
\multicolumn{1}{c}{3} & kidneys & csf & atrium & encephalopathy & hemoptysis & \textit{attacks} & \textit{sodium} & podiatry & qhs & dysfunction & pain/shortness & meropenem \\
\multicolumn{1}{c}{4} & baseline & ekg & emg & angiopathy & chemo & sclerosis & sat & opthamology & \textit{once/day} & function & \textit{breath} & daptomycin \\
\multicolumn{1}{c}{5} & renal & load & \textit{ecg} & metastatic & diverticulosis & endocarditis & 2l & \textit{eye} & po/ng & adrenal & fevers & \textit{antibiotics} \\ \hline
\multicolumn{13}{c}{\textit{Embeddings without Subword Information, with Anchors (Clinician-designed Word Pairs)}} \\ \hline
\multicolumn{1}{c}{1} & \textit{creatinine} & \textit{x-ray} & echocardiogram & \textit{liver} & \textit{cancer} & \textit{attack} & \textit{sodium} & \textit{eye} & \textit{daily} & \textit{kidney} & \textit{breath} & zosyn \\
\multicolumn{1}{c}{2} & potassium & \textit{xray} & cardiac & cirrhosis & metastatic & infarction & moniter & opthamology & qday & liver & shortness & ceftriaxone \\
\multicolumn{1}{c}{3} & renal & resolution & lab & organ & lymphoma & myocardial & bun & ophthalmologist & po & \textit{kidneys} & worsening & \textit{antibiotic} \\
\multicolumn{1}{c}{4} & rose & lungs & infarction & attempt & represent & arrest & potassium & podiatry & mg/day & disease & \textit{dyspnea} & cefepime \\
\multicolumn{1}{c}{5} & kidney & pleural & echo & portal & enlarged & nstemi & restriction & exam & twice & failure & palpitations & daptomycin \\ \hline
\multicolumn{13}{c}{\textit{Embeddings with Subword Information and MedlinePlus Augmentation, with Anchors (CHV Word Pairs)}} \\ \hline
\multicolumn{1}{c}{1} & renal & load & echocardiogram & pancreatic & malignant & \textit{attack} & \textit{sodium} & \textit{eye} & \textit{daily} & \textit{kidney} & shortness & \textit{antibiotic} \\
\multicolumn{1}{c}{2} & \textit{creatinine} & ekg & echocardiograms & pancreatitis & \textit{cancer} & infarct & gm & ophthalmologist & dailyyou & adrenal & shortnes & daptomycin \\
\multicolumn{1}{c}{3} & bun/creatinine & embolus & echocardiography & necrotic & polyps & atyou & bun & electrophysiology & coated & retinal & pain/shortness & ceftriaxone \\
\multicolumn{1}{c}{4} & potassium & edema & \textit{ecg} & lymphatic & carcinoma & myocardial & potassium & neurologists & isosorbide & \textit{kidneys} & lightheadedness & cefazolin \\
\multicolumn{1}{c}{5} & rapamune & xr & echo & pancreas & mass & mvp & serum & electrophysiologist & dailyplease & original & \textit{breath} & \textit{antibiotics} \\ \hline
\multicolumn{13}{c}{\textit{Embeddings without Using Anchors}} \\ \hline
\multicolumn{1}{c}{1} & report & made & relieved & operating & tell & status & sternal & tight & constipation & report & 08:30 & discharged \\
\multicolumn{1}{c}{2} & 100.5 & levofloxacin & continually & machinery & present & mental & temp & shown & quantity & notify & 09:40 & vital \\
\multicolumn{1}{c}{3} & greater & 750 & increasing & while & secretary & lethargy & pounds & reign & output & 0.1 & 10:00 & stablized \\
\multicolumn{1}{c}{4} & 101 & ciprofloxacin & headache & illicit & TRUE & confusion & 101.5make & frontal & redness & fo & 09:00 & haven \\
\multicolumn{1}{c}{5} & sharp & regimen & ha & drowsy & expiratory & hallucinations & 101.5 & lacerations & bloating & watch & 11:00 & supplemental \\ 
\bottomrule
\end{tabular}
}
\caption{Examples of the top-5 nearest neighbor words (identical word excluded) in the consumer language embedding space queried by professional words (first row). Many commonly used appropriate corresponding consumer words for each queried clinical professional word are seen in the list. Italicizing words represent functionally correct word translation.}
\label{tab:table5}
\vspace{-20pt}
\end{table*}

Subword embeddings utilize character $n$-gram information, which is useful in capturing lexical and morphological patterns. 
The retrieved words from embeddings with subword information tend to retrieve a group of morphologically similar words, e.g. ``mi'' $\rightarrow$ ``attack'' and ``attacks'' (Table~\ref{tab:table5}). 
However, the performance drops when the morphologically similar words are semantically incorrect, e.g. ``ophthalmology'' $\rightarrow$ ``ob-gyn'' and ``ob/gyn'', as in our case.
Instead, word embeddings without subword information focus more on whole-word semantics, such as synonyms and antonyms, with different lexical morphologies. 
This is one of the reasons why subword embeddings can't always yield better results even though they utilize more information.
Because they have different strengths in translation, we kept both for sentence translation. 

Although the 1000-dimension subword embedding with MedlinePlus augmentation yields superior performance in CHV pairs evaluation, we didn't see additional benefits in the qualitative evaluation compared to the 100-dimension version without augmentation, due to the tendency to retrieve professional-level words in the 1000-dimension version.
This is because the translated words in CHV pairs are not always in consumer-understandable language, such as ``abd'' $\rightarrow$ ``abdomen'', and therefore using the parameters based on the CHV pairs evaluation is not reliable.

We also conclude that the embeddings without using anchors during BDI yields the worst performance in word translation.
Using eigenvector score, we found that adopting anchors yielded higher embedding space similarity than without anchors. The eigenvector scores decrease 16.7\% ($0.54 \rightarrow 0.45$) in subword embeddings, and decreases 47.8\% ($0.46 \rightarrow 0.24$) in word embeddings after applying anchors.

For sentence translation, we decided to adopt (1) 100-dimension \texttt{MUSE}-aligned embedding spaces with subword information and without data augmentation, (2) 1000-dimension \texttt{MUSE}-aligned embedding spaces with subword information and with data augmentation, and (3) 300-dimension \texttt{MUSE}-aligned embedding spaces without subword information and without data augmentation.

\subsection{Sentence Translation}
In Table~\ref{tab:table6}, we evaluate the translation by correctness as judged by clinicians, and readability by both clinicians and non-clinicians.

\begin{table}[]
\begin{tabular}{cccc}
\toprule
\multicolumn{1}{c}{Configuration} & \begin{tabular}[c]{@{}c@{}}Correctness\\ (Clinicians)\end{tabular} & \begin{tabular}[c]{@{}c@{}}Readability\\ (Clinicians)\end{tabular} & \begin{tabular}[c]{@{}c@{}}Readability\\ (Non-clinicians)\end{tabular} \\ 
\midrule
\textit{A} & 2.85 (1.41) & 4.48 (0.84) & 4.02 (1.17) \\
\textit{B} & 2.89 (1.47) & 4.35 (0.70) & 3.60 (1.14) \\
\textit{C} & 2.95 (1.55) & 4.56 (0.75) & 4.26 (0.95) \\
\textit{D} & 3.33 (1.46) & 4.03 (1.00) & 3.81 (0.79) \\
\textit{E} & 3.57 (1.34) & 4.55 (0.65) & 4.28 (0.93) \\
\textit{F} & 4.10 (1.22) & 4.59 (0.65) & 4.25 (0.83) \\
\textit{N} & 1.18 (0.56) & - & - \\ 
\midrule
Dictionary-based & 4.13 (0.62) & 3.57 (1.08) & 3.55 (0.86) \\
\bottomrule
\end{tabular}
\caption{Performance of sentence translation using our unsupervised SMT framework. The values are the average (standard deviation) of mean opinion score (MOS) regarding the correctness and readability of translated sentences. The readability is accessed only for the sentences with correctness score $\geq$ 4. For configurations, please refer to Table~\ref{tab:table2}. Baseline is the supervised, dictionary-based CHV replacement method.}
\label{tab:table6}
\vspace{-20pt}
\end{table}

The supervised replacement baseline yields the best correctness score since it doesn't change the semantics of sentences too much.
However, its readability scores are lower.
This is because the CHV mapping doesn't align with the actual consumer language well, and still keeps many professional terms after replacement.

\paragraph{Selecting BDI-aligned Embedding Spaces}
Configuration $F$, which used the 300-dimension word embeddings with anchors in BDI and adopted the MIMIC-consumer corpus for language modeling, yields the highest correctness scores among all SMT configurations.
Followed by configuration $E$, $D$, $C$, $B$, $A$, then $N$, we found that the most critical component for sentence translation is using identical strings as anchors for BDI. 
Mere sentences are correct and no sentences reached the threshold for readability evaluation (correctness score $\ge 4$) if anchors were not used (configuration $N$).
This emphasizes the critical role of anchors in language translation without supervision (Table~\ref{tab:table5}).

Using word embeddings trained without subword information (configurations $E, F$) provided better correctness than those adopting subword information (configurations $A, B, C, D$).
The reason is similar to word translation---training with subword information captures more morphologically similar words, yet the performance drops when those morphologically similar words are clinically incorrect. 
Instead, using word information captures more synonyms. 
The morphological errors can be corrected while applying language models during sentence translation.

Even though the 1000-dimension subword embeddings with MedlinePlus augmentation (configurations $C, D$) outperformed other embeddings significantly on word-level CHV evaluation, its correctness on sentence translation is limited.
This provides evidence that CHV pairs are less aligned with clinical narratives than the clinician-designed word pairs, and therefore the optimal setting for CHV is not the same as for real clinical narratives.

\paragraph{Language Modeling}
Choosing the specific corpus (configurations $B, D, F$) for language modeling yielded better correctness but inferior readability than using the general corpus (configurations $A, C, E$).
Language models trained on general corpora tend to reshape professional words to more general terms and phrases.
For example, mapping from ``flagyl'' (name of a kind of antibiotic) to ``antibiotics'' leads to better readability.
Instead, using the more specific MIMIC corpus gives us more explanations of clinical professional terminologies. 
For example, ``r femoral line'' $\rightarrow$ ``right central line'' and ``catheter'' $\rightarrow$ ``foley catheter''.
It also provides better ability to expand the medical abbreviation to the completed word---which may also be helpful for the professional to consumer language translation. 
For instance, ``afib'' $\rightarrow$ ``atrial fibrillation'', ``ppi'' $\rightarrow$ ``pantoprazole'', ``o2 sat'' $\rightarrow$ ``oxygen saturation''.
Such word and phrase identification and appropriate replacement is a critical first step for professional-to-consumer translation, which is not seen in simply using dictionary-based replacement method.
Table~\ref{tab:table7} displays few examples of sentence translation using configuration~$F$.

\begin{table}[]
\footnotesize
\begin{tabularx}{8.5cm}{ScX}
\toprule
\textit{Original} & the patient had an o2 saturation in the 80s when they arrived , with a heart rate in the 130s , atrial fibrillation at that time and blood pressure with a systolic of 200 . \\
\textit{SMT} & the patient had an \textit{oxygen saturation} in the 80s when they came , with a fast heart rate in the , \underline{atrial fibrillation} at that time and with a \textit{systolic blood pressure} of 200 . \\ \midrule
\textit{Original} & food impaction within the esophagus . \\
\textit{SMT} & \textit{food stuck} in your esophagus . \\ \midrule
\textit{Original} & her diet was gradually advanced to puree consistency . \\
\textit{SMT} & your diet was gradually advanced to \textit{thickened liquids} . \\ \midrule
\textit{Original} & cxr from showed no infiltrate but did show volume overload . \\
\textit{SMT} & from \textit{x-ray} showed {no pneumonia} but did show volume overload . \\ \midrule
\textit{Original} & the patient is not on a statin because of hepatic dysfunction . \\
\textit{SMT} & the patient is not on a \underline{statin} because of \textit{liver function} . \\ \midrule
\textit{Original} & a bal was done on and was still showing yeast , by specimens were clear . \\
\textit{SMT} & a \textit{bronchoscopy} was done on and was still showed \underline{yeast} , by \textit{cultures} were clear . \\ \midrule
\textit{Original} & her fs levels were noted to be between 120-150 on tubefeeds and po intake . \\
\textit{SMT} & your \textit{glucose levels} were found to be between 140 \textit{on nutrition} and \underline{po} intake \\ \midrule
\textit{Original} & humalog iss discontinued on for good glycemic control . \\
\textit{SMT} & \textit{stopped on sliding scale for glucose control} good . \\ \midrule
\textit{Original} & pathology shows metastatic carcinoma morphologically consistent with breast origin , present at cauterized surgical margin . \\
\textit{SMT} & \textit{biopsy showed cancer morphologically consistent with breast cancer} , causing present at \textit{clipped surgical area} . \\ \midrule
\textit{Original} & r femoral line placed . \\
\textit{SMT} & \textit{right central} line placed . \\ \midrule
\textit{Original} & a cystoscopy was performed intraop and a catheter was placed . \\
\textit{SMT} & a \textit{procedure} was performed and a \underline{foley} catheter was placed . \\ \midrule
\textit{Original} & the patient developed an abdominal abscess . \\
\textit{SMT} & the patient had an abdominal \textit{infection} . \\ \midrule
\textit{Original} & he has advanced well with his feeding . \\
\textit{SMT} & you have \textit{tolerated well with your nutrition} . \\ \midrule
\textit{Original} & the patient was subsequently reintubated secondary to pulmonary compromise . \\
\textit{SMT} & the patient was \textit{later intubated due to pulmonary injury} . \\ \midrule
\textit{Original} & she was treated with po levofloxacin with plan for a 14 day course . \\
\textit{SMT} & you were treated with \textit{oral antibiotics} with plan for a 14 day course . \\ \midrule
\textit{Original} & likely etiology was hypotension leading to underperfusion of coronary artery and possible conduction system abnormalities . \\
\textit{SMT} & was likely due to \underline{transient} hypotension \textit{complication} of coronary artery and possible \textit{rhythm system} abnormalities . \\
\bottomrule
\end{tabularx}
\caption{Examples of unsupervised sentence translations. Italicizing words and phrases represent functionally correct translation, and underlining words and phrases represent inappropriate translation or the words/phrases require better translation.}
\label{tab:table7}
\vspace{-25pt}
\end{table}

Using different corpora for language modeling affects the quality of sentence translation, yet both general and specific corpora have their own limitations.
We already know that language models based on a general corpus help to convert the specific terms into general version.
Such language generalization is helpful for readability, yet it sometimes results in oversimplification, ambiguity and vagueness that misses the important information (e.g., ``hyponatremia'' $\rightarrow$ ``sodium'', ``troponin'' $\rightarrow$ ``cardiac''), and therefore reduces correctness.
In contrast, language models using a specific corpus can make better expansion of abbreviations and explanations. 
However, they may also make the translation be too specific, such as when ``vancomycin'' may be translated into ``flagyl'' or other antibiotics. 

Some corpus-specific words are usually replaced during the sentence translation, such as pronouns, commonly-seen dosages, medications and procedures.
For example, language models using a general corpus tend to translate ``pt'' to ``he'', yet using the MIMIC-consumer corpus tends to translate it into ``you'', which makes sense since the MIMIC-consumer corpus contains colloquial instructions written for patient and family.
Negation is also sometimes incorrect.
Abbreviation ambiguity is also an issue: e.g., ``pt'' can be translated to ``patient'', ``physical therapy'', or ``posterior tibial artery'', but the results may be decided by the language model we use.

Choosing the ideal corpus, which considering the trade-off between general and specific, to build language models for SMT is a critical step for deployment. Other techniques to explore include introducing a copy mechanism or using biomedical ontologies to identify words and phrases that should be fixed or preserved (e.g. medication and procedure names). We may also need techniques to handle pronouns and negations correctly.
For ambiguity, a contextualized word representation is an approach to be considered~\cite{peters2018deep}.

\section{Conclusions}
In this study, we recognize a strong need for clinical professional to consumer language translation, which is a difficult task even using the current state-of-the-art MT method and system from the general domain of NLP.
However, we demonstrate that our novel fully-unsupervised translation framework works in the setting of non-parallel corpora without the assistance of expert-curated knowledge, which is not possible using traditional approaches but essential for scalability while tackling real-world clinical data---since data availability and expert curation are always problems for clinical machine learning.
We utilized unsupervised natural language representations, iterative Procrustes process with anchor information for BDI, and SMT using language modeling and obtained promising performance in both word and sentence-level translations evaluated by two quantitative tasks, and validated by human (both clinicians and non-clinicians) judgment.
The newly-proposed two-step evaluation for sentence translation without ground truth reference, which considers both human experts (clinical correctness) and laymen (readability) judgements, is also helpful for practical use. 

Some limitations in our study shed light on future directions. 
With the proposed framework, we found that the readability of the translation is superior but the correctness is slightly inferior than using supervised dictionary-based word replacement, which mainly results from the trade-off between over-simplification and over-specification.
The loss of correctness or mistranslation in the clinical language translation may be harmful due to over-simplification. 

Possible solutions for such obstacle of deployment can be identified in two directions. 
First, we can leverage the current method by incorporating domain knowledge with the definition insertion from existing dictionaries.
We may use the clinical concept-level information by just focusing on translating the words and phrases that match UMLS concepts.
Linguistic features can also be considered in the process of translation.
The proposed machine learning-based approach can be combined with linguistic characteristics of the corpus~\cite{biran2011putting}, to have a better control of language simplification.
Quantitative evaluation that considering linguistic features may help strengthen the interpretability~\cite{feng2010comparison}.
Adopting contextualized representations is also an alternative to improve the quality of embeddings and models~\cite{peters2018deep}.
For generalizability, conducting experiments on datasets in other domains with similar settings is also considered.
Finally, we plan to expand of the clinician-designed word pairs set and deploy the framework for public use as an online translator.
The framework and results in this work could ultimately be useful to improve patient engagement in their own health care, and toward the era of personalized medicine.

\begin{acks}
The work is funded by MIT-IBM Watson AI Lab. The authors thank Dr. Tristan Naumann, Matthew McDermott, the MIT Clinical Decision Making Group, and all evaluators for their helpful discussions.
\end{acks}

\bibliographystyle{ACM-Reference-Format}
\balance
\bibliography{mybib}

\pagebreak


\appendix

\section{Supplemental Material}
In the supplement, we provide the details of model training and evaluation methods for reproducibility.
The scripts are available at the project repository (\texttt{https://github.com/ckbjimmy/p2c}).

\subsection{Word Translation}
\label{app:1}

\paragraph{Learning Word Embeddings}
We trained the word embeddings by setting the word window size $k=5$. We considered all words that appear more than once, with a negative sampling rate of $10^{-5}$. 
The model was trained by stochastic gradient descent~(SGD) without momentum with a fixed learning rate of $0.1$ for 5 epochs. 

To include the subword information, we considered the length of character n-grams between 2 and 5, and used the fastText implementation for word representation learning for all experiments. 
We experimented on an embedding dimension of 50, 100, 200, 300 and 500 for clinician-designed word pairs evaluation, and 100, 200, 300, 500, and 1000 for CHV pairs evaluation.

\paragraph{Bilingual Dictionary Induction}
When we used the identical character strings as anchors for unsupervised BDI, we did 5 iterations of Procrustes process for \texttt{MUSE}~\cite{conneau2018word}. 
We adopted the default setting for \texttt{VecMap} self-learning, which used symmetric re-weighting before and after applying SVD, and bidirectional dictionary induction~\cite{artetxe2018robust}. 

Without anchors, we utilized adversarial learning~\cite{conneau2018word}.
For the discriminator in adversarial training, we used a two-layer neural network of size 2048 without dropout, and leaky rectified linear unit (ReLU) as the activation function. 
We trained both the discriminator and~$W$ by stochastic gradient descent (SGD) with a decayed learning rate from 0.1 to $10^{-6}$ with the decay rate of 0.98. 
We selected the 1000 most frequent words for discrimination. 
For refinement, we also did 5 iterations of Procrustes process.

The evaluation pairs are available through the project repository.

\subsection{Sentence Translation}
\label{app:2}

\paragraph{Language Modeling}
In language modeling, we applied the default Moses smoothed $n$-gram language model with phrase reordering disabled during the very first generation~\cite{lample2018phrase,koehn2007moses}. 
We trained the model iteratively while randomly picking source sentences for translation. 
The length of phrase tables is 4. 
The hyperparameter $T$ for computing softmax scores is set to be 30.

\subsubsection{Evaluation of Sentence Translation}
\begin{figure}[hb]
\centering
\includegraphics[width=0.8\linewidth]{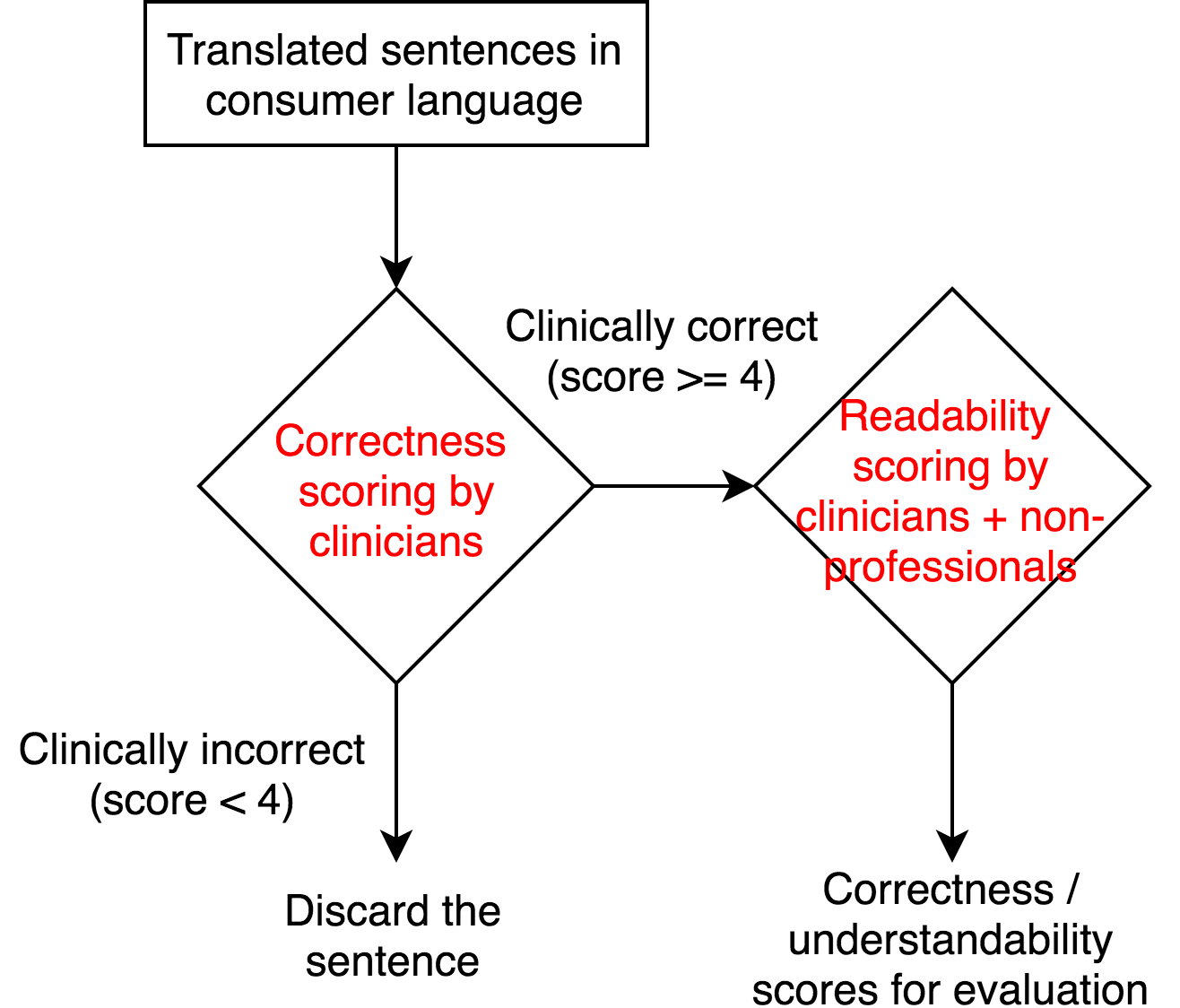} 
\hfill
\caption{Evaluation process of sentence translation.}
\label{fig:figS1}
\vspace{-10pt}
\end{figure}

In Figure~\ref{fig:figS1}, we demonstrate the workflow of the translated sentence evaluation. 

The two-step sentence evaluation considers clinical correctness and consumer (both clinicians and non-clinicians) readability.
We first asked clinicians to score the sentence correctness from the clinical perspective.
Criteria and examples for scoring correctness and readability are shown in Table~\ref{tab:tableS1}.

\begin{table*}[htbp]
\begin{tabularx}{\textwidth}{ScX}
\toprule
\multicolumn{1}{c}{Score} & \multicolumn{1}{c}{Criteria / Example} \\ \hline
\multicolumn{1}{c}{\textit{\textbf{}}} & \multicolumn{1}{c}{\textit{Correctness (consider both original and translated sentences)}} \\ \hline
\multicolumn{1}{c}{5} & 0 or 1 error. \\
\multicolumn{1}{l}{} & Original: chest tubes were discontinued without incident . \\
\multicolumn{1}{l}{} & Translation: chest tubes were stopped without complications . \\ \hline
\multicolumn{1}{c}{4} & 2 errors, or the same as professional language. \\
\multicolumn{1}{l}{} & Original: it was stable on monitoring , and her stools were guaiac negative . \\
\multicolumn{1}{l}{} & Translation: they were stable , and were monitoring the black blood was negative . \\ \hline
\multicolumn{1}{c}{3} & 3 errors, or with ambiguity that not easy to be inferred from clinical knowledge. \\
\multicolumn{1}{l}{} & Original: per family , she had + culture of a very resistant bacteria that is not mrsa . \\
\multicolumn{1}{l}{} & Translation: per family , you had no culture of a very different bacteria that is not cellulitis . \\ \hline
\multicolumn{1}{c}{2} & With the key clinical concepts but in the directions, numbers that don't make sense in clinical setting. \\
\multicolumn{1}{l}{} & Original: the patient had an o2 saturation in the 80s when they arrived , with a heart rate in the 130s , atrial fibrillation at that time and blood pressure with a systolic of 200 . \\
\multicolumn{1}{l}{} & Translation: the patient had an oxygen saturation in the 30 when they arrived to heart with a rate in the 30 to atrial fibrillation at that time and blood pressure with a systolic of 2000 . \\ \hline
\multicolumn{1}{c}{1} & Missing any critical information. \\
\multicolumn{1}{l}{} & Original: her wound was debrided at in the beginning of with several days of icu stay . \\
\multicolumn{1}{l}{} & Translation: the site was at the foot of the next few days for a hospital stay . \\ \hline
\multicolumn{1}{c}{\textit{\textbf{}}} & \multicolumn{1}{c}{\textit{Readability (consider only translated sentence)}} \\ \hline
\multicolumn{1}{c}{5} & 0 or only 1 word can't understand. \\
\multicolumn{1}{l}{} & Translation: a scan was performed which showed a right upper small clots . \\ \hline
\multicolumn{1}{c}{4} & 2 words. \\
\multicolumn{1}{l}{} & Translation: coumadin was held that night , and inr was therapeutic on . \\ \hline
\multicolumn{1}{c}{3} & 3 words, or confusing about wordings. \\
\multicolumn{1}{l}{} & Translation: you were given coumadin 5 mg po qam and your inr was monitored daily . \\ \hline
\multicolumn{1}{c}{2} & More than 4 words, or need explanations to understand. \\
\multicolumn{1}{l}{} & Translation: improved sat with improving ms . . \\ \hline
\multicolumn{1}{c}{1} & Completely can't understand, semantically meaningless. \\
\multicolumn{1}{l}{} & Translation: person to person you service them and the pressures and neurosurgery . \\
\bottomrule
\end{tabularx}
\caption{Criteria for correctness and readability scoring.}
\label{tab:tableS1}
\vspace{-10pt}
\end{table*}

For correctness, we consider negation error (yes/no, true/false, positive/negative), numeric/dosage error (e.g. 32 $\rightarrow$ 10), different named entities or semantics (e.g. ca $\rightarrow$ potassium, vancomycin $\rightarrow$ cefazolin, increase $\rightarrow$ decrease, hypertension $\rightarrow$ hypotension), or missingness of the critical information, as incorrectness.
The errors of ambiguities that don't change the understanding of sentence (e.g. hospital $\leftrightarrow$ icu, pathology $\leftrightarrow$ biopsy, some lung signs such as opacity (which can be detected in x-ray) $\leftrightarrow$ x-ray), pronouns (he, she, they, you), typos, duplicated words, punctuation, and slightly incorrect grammar are acceptable.
We only kept the sentences with a mean correctness score $\geq 4$ for readability evaluation.

For readability, we consider whether the sentence is understandable for consumers. 
Thus, we accept grammatical and syntactic errors that don't confuse evaluators (e.g. a bone scan showed no on cancer disease .), and no need to consider whether it is a reasonable clinical condition since this should already be judged by professionals while doing correctness scoring.
We also asked clinicians to score the readability, but the scores from non-clinicians and clinicians are calculated separately.

Before assigning the original and translated sentences, we roughly filtered out the sentence sets with severe incompleteness, format errors (e.g. too many numbers, duplicated words, punctuation errors), or severe fragmented sentences. 
Minimal format errors are acceptable during evaluation. 
The sentences related to neonatal patients were excluded since they are very specific and we expect that the data distribution is very different from the sentences describing adult patients. 
Finally, we assigned 1000 translated sentences to 20 evaluators (10 clinical and 10 non-clinical).

\end{document}